\pdfoutput=1

\documentclass[11pt]{article}

\usepackage{acl}

\usepackage{times}
\usepackage{latexsym}
\usepackage{graphicx}
\usepackage{caption}
\usepackage{subcaption}
\usepackage{multirow}
\usepackage{longtable}
\usepackage{todonotes}
\usepackage[T1]{fontenc}
\usepackage[utf8]{inputenc}
\usepackage{microtype}
\usepackage{inconsolata}

\title{The Curious Case of Representational Alignment: Unravelling Visio-Linguistic Tasks in Emergent Communication}

\author{Tom Kouwenhoven \\ LIACS \\ Leiden University \\ \texttt{t.kouwenhoven@liacs.leidenuniv.nl}
        \And
        Max Peeperkorn \\ School of Computing \\ University of Kent \\ \texttt{m.peeperkorn@kent.ac.uk} 
        \AND
        Bram van Dijk \\ LUMC \\ Leiden University Medical Center \\ \texttt{b.m.a.van\_dijk@lumc.nl} 
        \And 
        Tessa Verhoef \\ LIACS \\ Leiden University \\ \texttt{t.verhoef@liacs.leidenuniv.nl}}

\begin{document}
\maketitle
\begin{abstract}
Natural language has the universal properties of being compositional and grounded in reality. The emergence of linguistic properties is often investigated through simulations of emergent communication in referential games. However, these experiments have yielded mixed results compared to similar experiments addressing linguistic properties of human language. Here we address \textit{representational alignment} as a potential contributing factor to these results. Specifically, we assess the representational alignment between agent image representations and between agent representations and input images. Doing so, we confirm that the emergent language does not appear to encode human-like conceptual visual features, since agent image representations drift away from inputs whilst inter-agent alignment increases. We moreover identify a strong relationship between inter-agent alignment and topographic similarity, a common metric for compositionality, and address its consequences. To address these issues, we introduce an alignment penalty that prevents representational drift but interestingly does not improve performance on a compositional discrimination task. Together, our findings emphasise the key role representational alignment plays in simulations of language emergence.
\end{abstract}

\section{Introduction}
Human language bears unique properties that make it a powerful tool for communication. A well-known property is compositionality: the ability to combine meaningful words into more complex meanings \citep{hockett1959animal}. The emergence of compositionality is studied extensively in the field of language evolution through human experiments \cite[e.g.][]{Selten2007Emergence, Kirby2008CumulativeLanguage, Kirby2015CompressionStructure, Raviv2019CompositionalTransmission}. An important finding from this field is that the unique nature of human language can be explained as a consequence of biases for simplicity and expressivity imposed during continuous language learning and use \citep{Smith2022HowStructure}. Computational simulations of language emergence have also been used to study the emergence of linguistic properties \cite[e.g.][]{DeBoer2006ComputerEvolution, Steels2012TheGame}, and have seen a rising interest in the field of computational linguistics \citep{lazaridou2020emergent}. Here, compositionality in the emergent communication protocols is commonly measured through topographic similarity \cite[\textsc{topsim};][]{Henry2006Understanding}. This metric was first used in recent computational simulations by \citet{Lazaridou2018EmergenceInput} and has been used in a large body of work since. Yet, the interpretation of linguistic properties emerging in simulations remains challenging, since language protocols used among artificial agents often show critical mismatches with known properties of human languages \citep{galke2022emergent, lian2023communication} such as efficiency, word-order vs. case-marking biases, or compositional generalisation (see §\ref{background}). Consequently, it is evident that their learning biases and signal-meaning mappings differ from those of humans. This underscores the critical need to obtain deeper insight into referential games in the language learning setting \citep{rita2022emergent}.

A possible explanation for these mismatches could stem from representational alignment, the degree of agreement between the internal representations of two information processing systems \citep{sucholutsky2023getting}. To the best of our knowledge, representational alignment in emergent communication was first reported by \cite{bouchacourt-baroni-2018-agents}, who measured the degree to which agents aligned their internal image interpretations (inter-agent alignment) by performing Representational Similarity Analysis \cite[\textsc{rsa}; ][]{kriegeskorte2008representational}. Using \textsc{rsa} (§\ref{sec:alignment}), they showed that agents establish successful communication artificially by aligning their internal image representations while \textit{losing} any relation to the images presented (image-agent alignment), enabling communication about noise input even though they were trained on real images. As such, their communication protocol captured not conceptual properties of the objects depicted in pictures, but most likely focused on non-human-like spurious image features (e.g., pixel intensities). While inter-agent alignment is not a problem per se, the loss of image-agent alignment is problematic for two reasons. First, for emergent communication simulations to provide meaningful insights into the emergence of natural human language, agent image representations must be grounded in the content of the images. Only then can we deduce \textit{what} the agents communicate about and assess linguistic properties or their ability to generalise to novel concepts. Second, emergent communication setups have been proposed to fine-tune pre-trained (vision-)language models, aiming to enhance machine understanding of natural human language \citep{lazaridou2020emergent, Lowe2020On, steinertthrelkeld2022emergent, zheng2024iterated}. In this context, maintaining substantial alignment between representations and images is crucial for preserving mutual understanding between machines and humans.

Representational alignment, however, did not receive the necessary attention since a host of papers appeared \textit{after} the findings by \citeauthor{bouchacourt-baroni-2018-agents} in which results on referential games were reported without taking \textsc{rsa} into account \citep[e.g.][]{Lazaridou2018EmergenceInput, guo2019emergence, LiBowling2019Ease-of-Teaching, Ren2020Compositional, chaabouni-etal-2020-compositionality, dagan-etal-2021-co, Mu2021Emergent, chaabouni2022emergent}. Admittedly, some use attribute-value objects instead of real images as input. But importantly, in nearly all cases, neural agents must map inputs—whether attribute-value objects or image representations—onto agent-specific representations. Therefore the problem of inter-agent alignment \textit{can always} occur and is \textit{agnostic} to the input type. Although this warrants further analysis of earlier results, the field is already employing referential games in more complex simulations with real images \cite[e.g.][]{Roberto2021Interpretable, chaabouni2022emergent, Mahaut2024ReferentialNetworksHeterogeneous}. 

\begin{figure*}[t]
    \centering
    \begin{subfigure}[t]{0.32\textwidth}
        \centering
        \includegraphics[width=\textwidth]{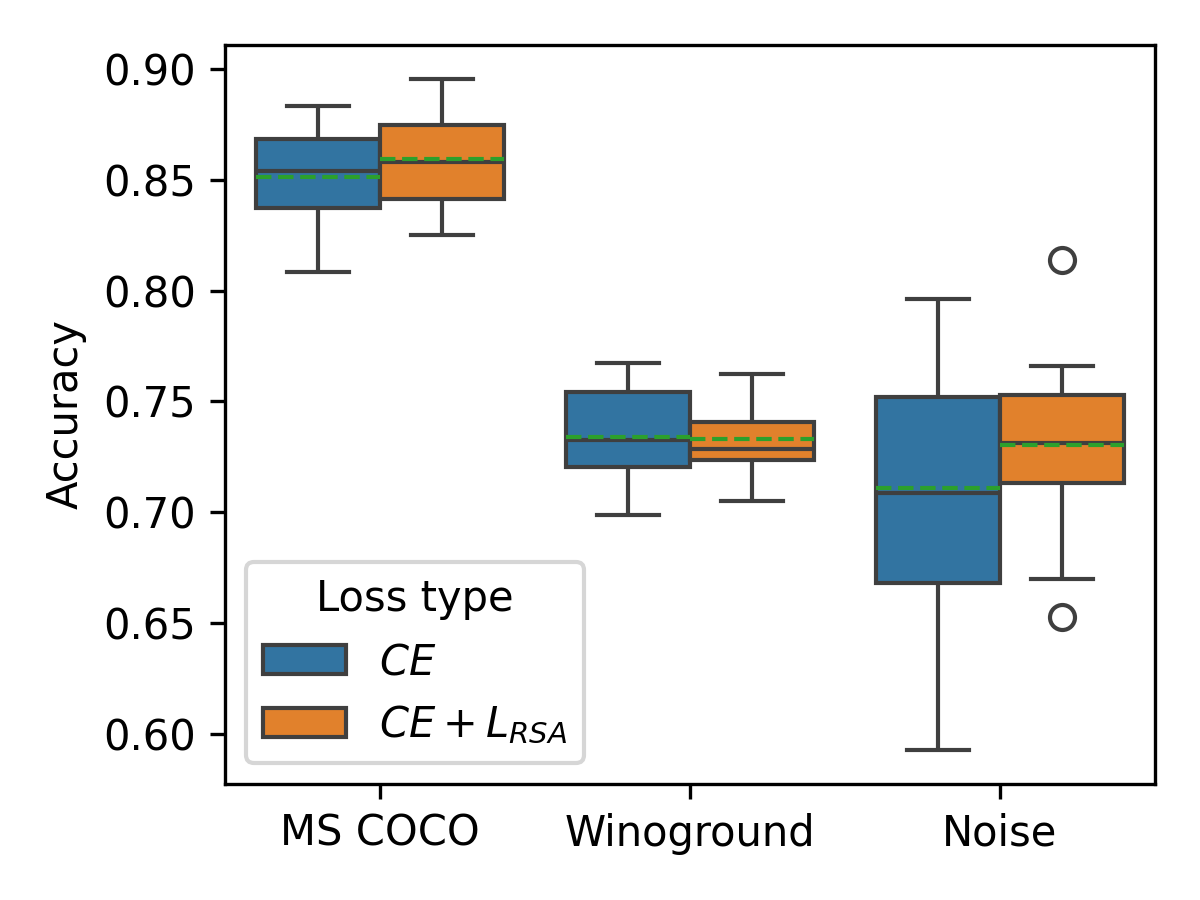}
        \caption{Communicative performance (Accuracy) on discriminating two images.}
        \label{fig:acc}
    \end{subfigure} 
    \hfill
    \begin{subfigure}[t]{0.32\textwidth}
        \centering
        \includegraphics[width=\textwidth]{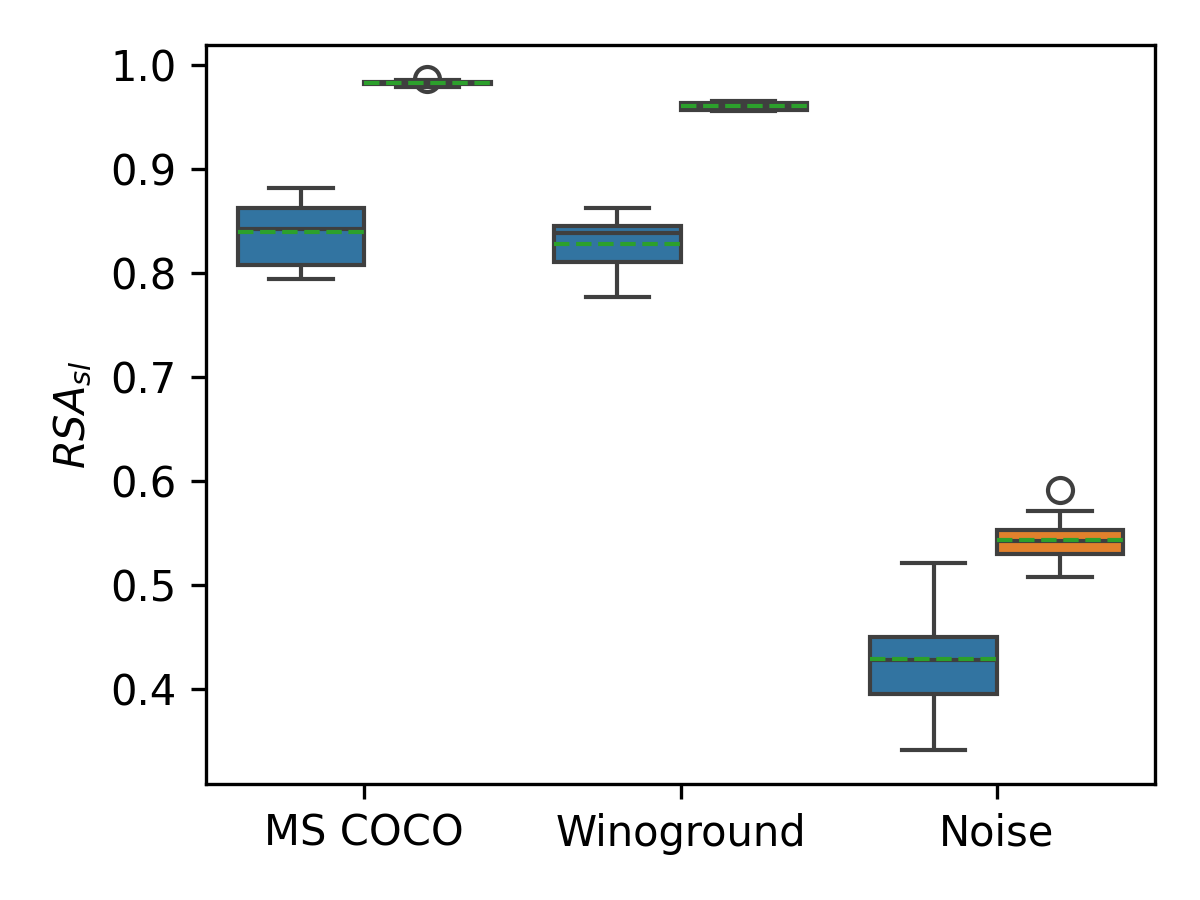}
        \caption{Inter-agent representational alignment ($\textsc{rsa}_{sl}$) between agent representations.}
        \label{fig:rsa_sl}
    \end{subfigure}
    \hfill
    \begin{subfigure}[t]{0.32\textwidth}
        \centering
        \includegraphics[width=\textwidth]{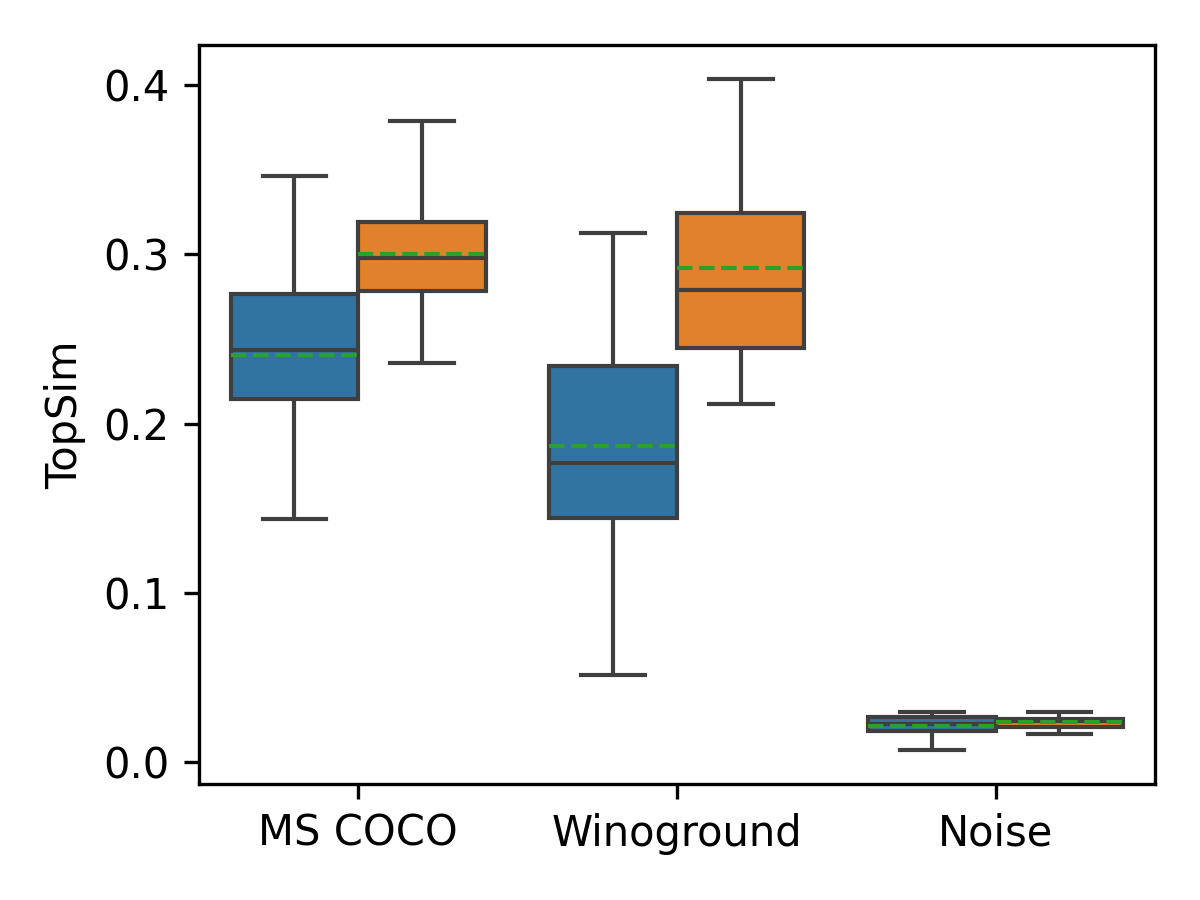}
        \caption{Topographic similarity (\textsc{topsim}) between the images and the messages.}
        \label{fig:topsim}
    \end{subfigure}
    \caption{Inference results for different datasets after training on MS COCO images. In (a) we see that agents can discriminate MS COCO images but struggle with discriminating Winoground images. In (b) we see the effect of the loss function on the degree of inter-agent representational alignment and (c) implies that according to the \textsc{topsim} metric, messages are more structured if the alignment penalty is used. The presented results are across 15 seeds and use the best-performing parameters resulting from our parameter sweep, dashed green lines indicate averages.}
    \label{fig:overall-results}
\end{figure*}

This work addresses the understudied alignment problem in standard referential game setups used in emergent communication. We train Reinforcement Learning (RL) agents equipped with a recent vision module \cite[DinoV2;][]{oquab2024dinov} to communicate about images. In addition to evaluating the agents on MS COCO \citep{lin2014microsoft} image pairs, we evaluate on noise pairs and image pairs sourced from the Winoground dataset \citep{Thrush2022Winoground}. The latter is explicitly created to gauge visio-linguistic compositional reasoning abilities of vision and language models. We first confirm that effective communication in the referential game relies on inter-agent alignment and then move on to our contributions. First, we find a strong correlation between the degree of inter-agent alignment and the \textsc{topsim} metric. Our second contribution consists of a solution to the alignment problem by including an alignment penalty term to the loss, resulting in equivalent communicative success and higher \textsc{topsim} whilst ensuring that the agents communicate about images instead of spurious features (Figure \ref{fig:overall-results}). We then argue to start evaluating emergent communication protocols on more strict tasks that directly target the intuition behind popular metrics to obtain a clearer understanding of the protocols. Overall, our results highlight the importance of representational alignment in simulations of language emergence and underscore the need to better understand the divergence in human and artificial language emergence.

\section{Background}
\label{background}
Most research in simulating emergent communication is modelled after the Lewis signalling game \citep{lewis1969convention} with a speaker and a listener agent. The speaker observes a state (e.g., an image) and sends a signal to the listener who acts based on this signal. In the case of the referential game, this means selecting a target among distractors. Both agents are rewarded for successful communication, meaning the listener points to the target object. The solution of this game requires the agents to have a shared protocol (i.e., an artificial language) which typically emerges when the agents learn based on trial and error over multiple games. This resembles how for humans, language learning and use impose constraints like pressures for learnability and compression that shape our language design \citep{Kirby2014IteratedLanguage, Kirby2015CompressionStructure}. Importantly, the emergent language in this setup is also shaped by biases resulting from, for example, the agent architecture, loss function, and learning protocol \cite{rita2022emergent}. The current work uses the referential game: a variant of the Lewis signalling game extensively used to explore language evolution \citep[e.g.][]{Steels2012TheGame, Kirby2015CompressionStructure, Lazaridou2017Multi-agentLanguage, kottur-etal-2017-natural, Lazaridou2018EmergenceInput, Kharitonov2020Entropy, chaabouni2022emergent}.

An important challenge in emergent communication is that artificial learners often do not behave the same way as human learners in experimental settings. Some emergent protocols do not follow Zipf's law and thus are anti-efficient unless pressures for brevity are introduced \citep{Chaabouni2019anti}, others do not show the word-order vs. case-marking trade-off found in human languages \citep{chaabouni-etal-2019-word, lian-etal-2021-effect}. Additionally, there is an ongoing debate on the degree to which the emergent languages allow for compositional generalisation \cite{lazaridou2020emergent, conklin2023compositionality}. It has been suggested to introduce communicative (e.g., alternating speaker/listener roles) and cognitive (e.g., memory) constraints \citep{galke2022emergent} and use more natural settings to promote more human-like patterns of language emergence with neural agents \citep{Kouwenhoven2022EmergingEvolution}. Doing so changes the learning pressures to which the agents need to adapt and can recover initially absent linguistic phenomena of natural language in emergent languages \cite[for a review see][]{galke2024emergent}. An example of such work, investigating the word-order vs. case-marking trade-off, has succeeded in replicating this trade-off for neural learners \citep{lian2023communication}. Their setup differs from other work in that agents first learn a miniature language via supervised learning, and then optimise it for communicative success via RL, resulting in emergent languages that share linguistic universals with human language. 

To enhance understanding of emergent communication in the Lewis game, \citet{rita2022emergent} decomposed the standard objective in Lewis games into two key components: a co-adaptation loss and an information loss. In doing so, they shed light on potential sources of overfitting and how they might hinder the emergence of structured communication protocols. They demonstrated that desired linguistic properties (e.g., compositionality and generalisability) emerge when they control the listener's ability to converge to the speaker agent (i.e., control for overfitting on the co-adaptation loss). While the co-adaptation loss has parallels to inter-agent alignment, their work does not address the alignment between the agents' image representation and the input features, which we deem crucial in developing grounded communication protocols.

Another challenge in emergent communication is the disentanglement of the underlying meanings of emergent languages. Earlier studies by \citet{Lazaridou2017Multi-agentLanguage} suggested that agents assign symbols to general conceptual properties of objects in images, rather than low-level visual features. However, as previously mentioned, follow-up work from \citet{bouchacourt-baroni-2018-agents} showed this is not always the case. They found that agents align their agent-specific image representations without developing a language that captures conceptual properties depicted in the images. Moreover, agents lost any sense of meaningful within-category variation where two similar objects in human perception (e.g., two avocados) were observed as maximally dissimilar for the agents. In response to these findings, recent studies have implemented sanity checks, testing whether trained agents can communicate about noise \cite{Roberto2021Interpretable, Mahaut2024ReferentialNetworksHeterogeneous}. However, to the best of our knowledge, there has been little attention to what we consider to be their main result: the alignment problem.  

\section{Representational alignment}
\label{sec:alignment}
Representational alignment is the degree of agreement between the internal representations of two information processing systems, whether biological or artificial. Even though widely recognised in cognitive science, neuroscience, and machine learning \citep{sucholutsky2023getting}, representational alignment has not seen much interest in the field of emergent communication, except for the work by \citeauthor{bouchacourt-baroni-2018-agents} who analysed the referential game using \textsc{rsa}. This metric measures the alignment between two sets of numerical vectors, for example, image embeddings and agents' representations thereof. In practice, it is calculated by taking the pairwise (cosine) distances between vectors of a set and calculating the Spearman rank correlation between these distances. 

In this paper, we also use \textsc{rsa} to operationalise representational alignment. Given the speaker image representations $r_{s}$ of the DinoV2 input embeddings $i$ and $r_{l}$ as the same images represented in the listener representation space, we compute the pairwise cosine similarity between the representations for the speaker $s_{s}$ and for the listener $s_{l}$ and calculate Spearman's $\rho$ between $s_{s}$ and $s_{l}$. As such, this measures the degree of inter-agent alignment ($\textsc{rsa}_{sl}$) between image representations $s_{s}$ and $s_{l}$, relative to their input. 
Additionally, we use it to measure image-agent alignment between the speaker/listener image representations and the DinoV2 embeddings ($\textsc{rsa}_{si}$ and $\textsc{rsa}_{li}$). Importantly, alignment is \textit{agnostic to the type of input}, being either images or attribute-value objects and can always happen when inputs are projected onto agent-specific representations. 

Intuitively, a high inter-agent $\textsc{rsa}_{sl}$ value can be interpreted as agents with \textit{similar} representations for similar images. Importantly, this can have two causes: both agents' image representations either \textit{maintain} a relation to the image input, or \textit{lose} this relation. While the former is desirable, the latter means that the agents are not communicating about the same high-level image features but are likely communicating about non-human-like spurious features. A low $\textsc{rsa}_{sl}$ value entails that the agents have developed \textit{different} interpretations for the same image. While this may well be similar to the question of whether people have different perceptual experiences of colour \citep{locke1847essay}, in the case of emergent communication, the agents should develop a grounded vocabulary with overlapping concept-level properties if we wish machines to have more natural understanding of human language. We use \textsc{rsa} 1) as a metric to re-assess findings from \citeauthor{bouchacourt-baroni-2018-agents} and 2) as an auxiliary loss to mitigate the alignment problem and ensure that the agents communicate about image features.   

\section{Methods}
The standard referential game is used as provided by the EGG framework \citep{kharitonovetal2021}. Doing so ensures our findings are representative of the widely-used setup, rather than being influenced by specific design decisions. The game is implemented as a multi-agent cooperative RL problem where a speaker and a listener communicate to discriminate a target image from two shuffled distractor images. The speaker receives a target $t$ and generates a message $m$ of at most length $L$, using vocabulary $V$. Importantly, the messages and symbols have no a priori meaning but are assumed to obtain meaning and become grounded during the game. Once meaningful, the symbols are ideally combined in a structured manner to create compositional messages that express more complex meanings. Using message $m$, the listener guesses the target $\hat{t}$. Communicative success is defined as $\hat{t} = t$, meaning that the listener has correctly identified the target image among the candidate images.

\subsection{Agents}
Agents contain a language and a vision module. The latter consists of a frozen pre-trained visual network (DinoV2) and a learned agent-specific representation layer. While difficult to know what conceptual image features are present in DinoV2 embeddings, they have demonstrated capability in semantic segmentation tasks \citep{oquab2024dinov}, which is similar to the agents' objective. In contrast to the hybrid structure of the vision module, the language module is entirely trained from scratch.

\textit{The speaker} agent processes images by applying a linear transformation to the image embeddings, followed by batch normalisation, to create its agent-specific image representation $r_s$. Its language module embeds this representation and passes it through a single-layer Gated Recurrent Unit \cite[GRU;][]{cho-etal-2014-learning} that spells out messages to describe the target. \textit{The listener} receives the message and the distractor images. It encodes the message into an embedding using another single-cell GRU layer. Additionally, a listener image representation $r_{l}$ is obtained for each image by applying a linear transformation and batch normalisation. Subsequently, temperature-weighted (temperature defaults to 0.1) cosine scores construct a multi-modal representation between the image and message representation \citep{Roberto2021Interpretable}, where a higher probability should be assigned to the target image.

\subsection{Optimisation}
Communicative success (\(\hat{t} = t\)) is used to optimise the trainable parameters of both agents. The listener minimises cross-entropy ($ce$) loss using stochastic gradient descent, amounting to supervised learning. The $ce$ loss is calculated over the listeners' target distribution, thus providing direct pressure for communicative success. At inference, the candidate image with the highest probability is chosen as the target $\hat{t}$. The gradients required to optimise the speaker are calculated using the \texttt{REINFORCE} \citep{williams1992simple} update rule as each generated symbol must be assigned a loss. Following common practice \citep{rita2024language}, entropy regularisation \citep{Mnih2016Asynchronous} is added to the loss to maintain exploration in message generation. 

In addition to the conventional $ce$ loss, we introduce an alignment loss ($ce+\textsc{rsa}$) that includes an alignment penalty term to enforce high inter-agent and image-agent alignment. The term \[L_{\textsc{rsa}} = (1-\textsc{rsa}_{sl})+(1-\textsc{rsa}_{si})+(1-\textsc{rsa}_{li})\] is added to the $ce$ loss with equal importance. We use torchsort \citep{blondel2020fast} to calculate $L_{\textsc{rsa}}$ such that the entire loss term is differentiable. Importantly, $L_{\textsc{rsa}}$ is not influenced by communicative success and does not interact with the $ce$ loss (Appendix \ref{sec:interaction_ce_ce_aux}). Only adding $\textsc{rsa}_{sl}$ to the $ce$ loss is not sufficient as high inter-agent alignment can be achieved while \textit{losing} image-agent alignment (see §\ref{sec:alignment}). We therefore also include $\textsc{rsa}_{si}$ and $\textsc{rsa}_{li}$ to ensure that the agents communicate about the content displayed in the images. Including $\textsc{rsa}_{sl}$ entails that representational information is shared between the agents, thus differing from how humans interact. Yet, ranking the speaker and listener representations in calculating $\textsc{rsa}_{sl}$ bears \textit{some} resemblance to projecting beliefs upon the interpretations of the other communicative partner. The current solution should be seen as a step towards more grounded vocabularies prone to refinements such as cognitive plausibility. We train for 30 epochs regardless of the loss used. The hyperparameters (Appendix \ref{sec:hyperparams}) that resulted in the best validation accuracy across 42 different communication channel capacities (Appendix \ref{sec:Appendix-capacity}) were used for our findings.

\subsection{Data}
\begin{figure}
    \centering
    \includegraphics[width=\columnwidth]{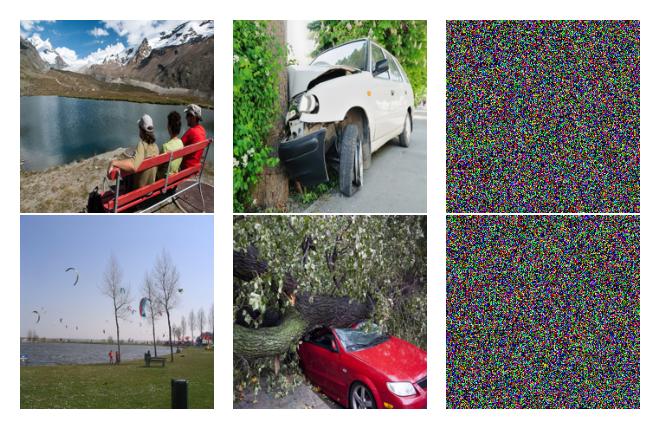}
    \caption{Exemplar pairs of each dataset used for evaluation. Left: an image pair from MS COCO. Middle: A Winoground example. Right: A Gaussian noise pair. All images are cropped for display purposes.}
    \label{fig:examplars}
\end{figure}

Agents are trained to discriminate MS COCO images but tested on three different datasets (Figure \ref{fig:examplars}) to assess out-of-distribution (o.o.d.) performance.

\textbf{MS COCO} --
We use a subset of 1200 images from the MS COCO 2017 validation set to train and test the agents using an 80/20 split. To obtain this subset, we first select the categories that contain more than 100 images (12 categories) and subsequently sample 100 images for each supercategory present in the resulting set of images. The distractor images are sampled from the same category to ensure that there is \textit{some} relevance to the target image. Importantly, sampling distractor images is done for each batch, meaning targets have different distractors at each epoch.

\begin{figure}[!ht]
    \centering
    \begin{subfigure}[t]{\columnwidth}
        \centering
        \includegraphics[width=\columnwidth]{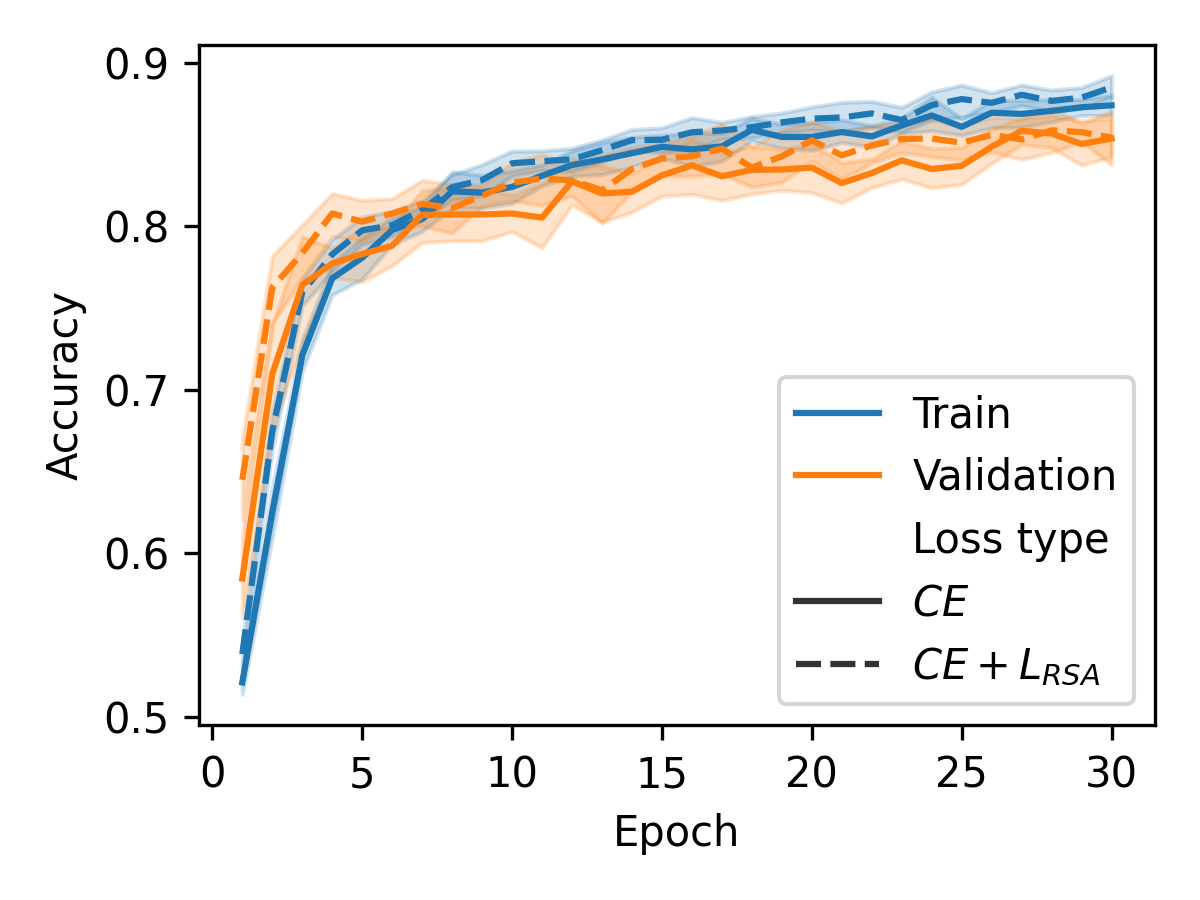}
        \caption{Learning curves for the MS COCO dataset on train and validation data.}
        \label{fig:evolution_acc}
    \end{subfigure} 
    \vfill
    \begin{subfigure}[t]{\columnwidth}
        \centering
        \includegraphics[width=\columnwidth]{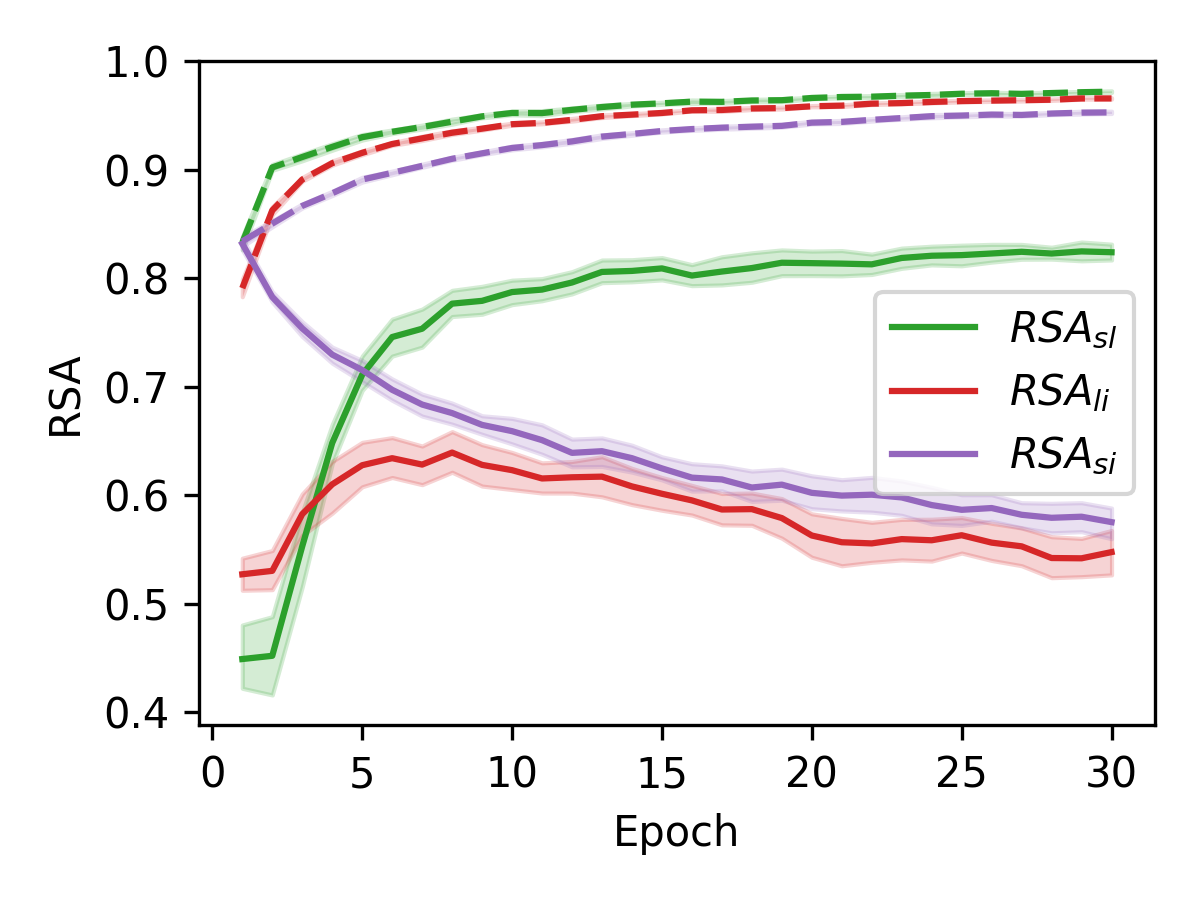}
        \caption{Representational alignment between agent image representations (green) and between the image and the sender/listener representations (purple, red).}
        \label{fig:evolution_rsa}
    \end{subfigure}
    \caption{In (a) we see that the agents learn to communicate successfully without overfitting on train data. In (b) we see that the alignment problem occurs with the $ce$ but not the $ce+\textsc{rsa}$ loss. Line style indicates the loss type. Data is averaged over 15 seeds, areas indicate the 95\% confidence intervals.}
    \label{fig:rsa_figure}
\end{figure} 

\textbf{Winoground} --
The Winoground dataset \citep{Thrush2022Winoground} was created to assess the visio-linguistic compositional reasoning abilities of vision and language models. Here, we repurpose it as a proxy for the agents' ability to endow in compositional reasoning for image-based settings. The dataset contains 800 images and corresponding captions, comprising 400 Winoground pairs. Image-caption pairs were included when the captions share the same words but are of different \textit{compositions}, implying completely different semantics (e.g., “a tree smashed into a car” versus “a car smashed into a tree” in Figure \ref{fig:examplars} (middle)). 
We only use the image pairs, not the captions. Crucially, this task differs from MS COCO since the image pairs are \textit{fixed}, \textit{conceptually similar} and meant to be discriminative if the agents' language allows for compositional reasoning and is grounded in the visual modality.

\textbf{Noise} --
Following \citet{bouchacourt-baroni-2018-agents}, we test whether agents can communicate about Gaussian noise ($\mu=0, \sigma=1$) pairs when trained on real images. Being able to do so would imply that messages communicate about spurious instead of high-level concept features.

\subsection{Metrics}
\label{sec:metrics}
The performance of our agents is assessed by communicative success (accuracy) and \textsc{rsa} (§\ref{sec:alignment}) measures alignment. The degree of compositionality in the emergent language is assessed through the \textsc{topsim} metric. Other metrics for compositionality like positional disentanglement, bag-of-symbols disentanglement \citep{chaabouni-etal-2020-compositionality} are not straightforward due to the continuous nature of the image embeddings. 

\section{Results}

\subsection{Communicative success}
\label{sec:communicative_success}
Unsurprisingly, results show that agents can successfully disambiguate between image pairs from MS COCO using an emergent language (Figure \ref{fig:evolution_acc}). Notably, we also confirm previous observations by \citep{bouchacourt-baroni-2018-agents} that agents trained on real images can communicate about Gaussian noise (Figure \ref{fig:acc}). Thus again suggesting that the messages convey spurious features rather than concept-level information. Interestingly, their performance on Gaussian noise is comparable to the performance on Winoground pairs, which requires the messages to capture concept-level properties. Thus revealing the difficulty of discriminating between strict pairs of conceptually similar images. The observed decrease in o.o.d. performance aligns with findings from other studies, such as \citet{Lazaridou2018EmergenceInput} and \citet{conklin2023compositionality}.

\subsection{The alignment problem}
\label{sec:alignment_problem}
The solid lines in Figure \ref{fig:evolution_rsa} clearly show that inter-agent alignment increases while alignment sensitivity to image features decreases for both agents. In principle, it is not a problem that the agents' image representations align. However, it is problematic when the alignment between the image embeddings and the image representations declines. Ablations across different channel capacities (§\ref{sec:Appendix-capacity}) and pre-trained vision modules (§\ref{sec:vision_modules}) showed that these trends appear consistently and are not influenced by the capacity or type of vision model. In addition to the communicative success on Gaussian noise, this re-confirms that the agents do not learn to extract concept-level information from the image embeddings but instead solve this task differently.

\subsection{\textsc{topsim} and representational alignment}
\label{sec:alignment_topsim}

\begin{figure}[!ht]
    \centering
    \includegraphics[width=\columnwidth]{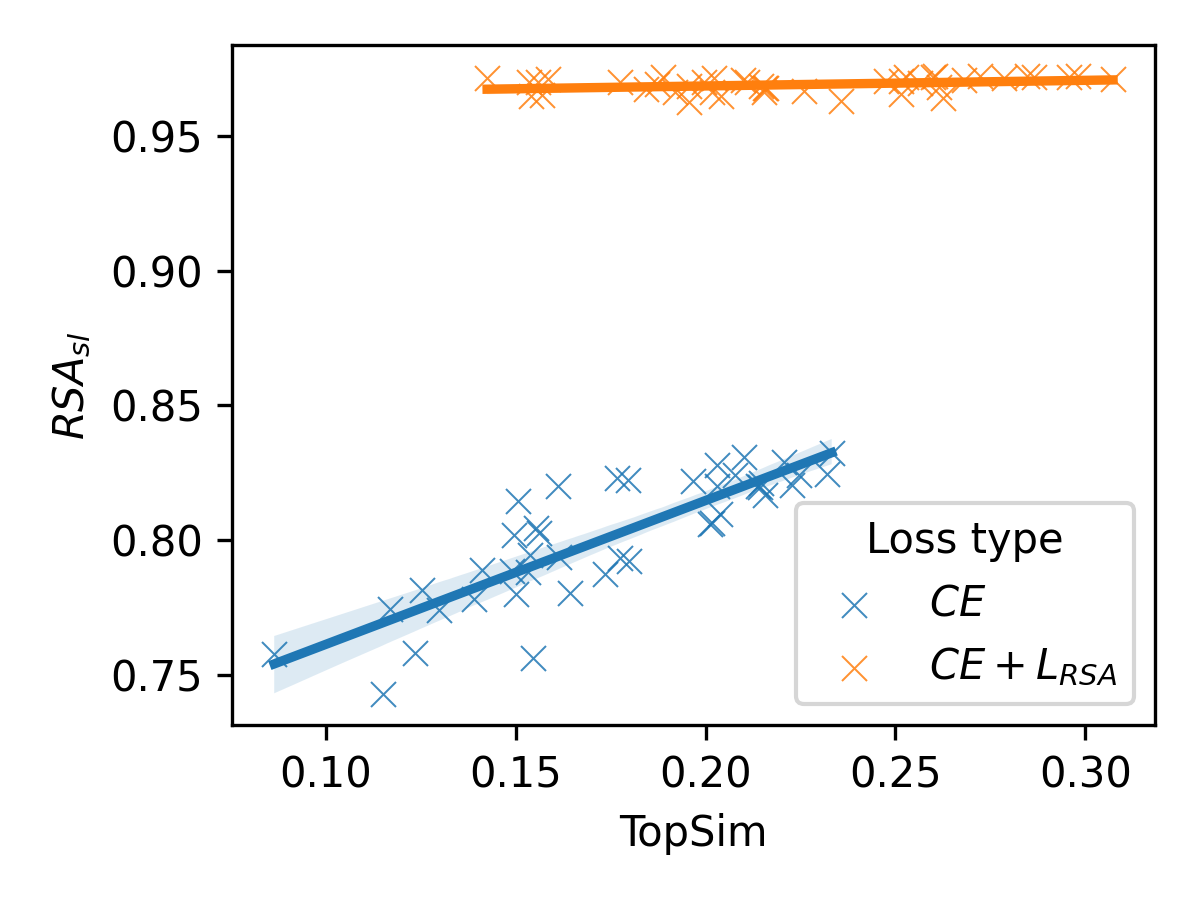}
    \caption{The relationship between \textsc{topsim} and inter-agent alignment ($\textsc{rsa}_{sl}$) for both loss types.}
    \label{fig:topsim-RSA}
\end{figure}

Earlier findings show mixed results on the relationship between \textsc{topsim} and generalisation in image-based settings, \textsc{topsim} was either related to generalisation \citep{chaabouni2022emergent} or not \citep{rita2022emergent}. Our results indicate that generalisation and \textsc{topsim} are correlated with both $ce$ (${r=.856}$, ${p<.001}$) and $ce+\textsc{rsa}$ (${r=.767}$, ${p<.001}$) losses. Meaning that more structured languages enable better communication on unseen validation pairs. Moreover, we find a strong positive relationship between $\textsc{rsa}_{sl}$ and \textsc{topsim} (${r=.838}$, ${p<.001}$) in the $ce$ (Figure \ref{fig:topsim-RSA}). This relation is also present in the $ce+\textsc{rsa}$ setup (${r=.408}$, ${p=.001}$), but is decoupled from \textsc{topsim} given the (very) small spread (${\sigma=.003}$) of $\textsc{rsa}_{sl}$. We do not observe an influence of inter-agent alignment on the number of uniquely produced messages.

\subsection{Mitigating the alignment problem}
\label{sec:alignment_mitigated}
We now focus on the $ce+\textsc{rsa}$ setup which was introduced to ensure that the agents maintain alignment with the image embeddings. Figure \ref{fig:evolution_rsa} shows that this is the case: inter-agent alignment \textit{and} agent-image alignment increase during training and remain high at inference. However, there does not seem to be a benefit for communicative success at inference time (Figure \ref{fig:overall-results}). This is because the alignment penalty only forces agents to represent images similarly to the image embeddings and is independent of the cross-entropy loss used to assess the success of communication (Appendix \ref{sec:interaction_ce_ce_aux}). In the case of noise images, we still observe the above-chance performance, suggesting that communication between the agents still occurs in an artificial manner. 

The alignment penalty also leads to increased \textsc{topsim}, indicating a higher level of structure (Figure \ref{fig:topsim}) and strengthens our finding that \textsc{topsim} and inter-agent alignment are related. Suggesting that the observed variations in \textsc{topsim}, whether higher or lower, as noted in previous studies \cite[e.g.][]{kottur-etal-2017-natural, chaabouni-etal-2020-compositionality}, should not be interpreted without considering alignment since they may be attributable to this underlying artefact rather than alterations to the original setup. 

When tested on more strict Winoground pairs, communicative success does not improve as a result of using the alignment penalty (Figure \ref{fig:acc}). Given the correlation between \textsc{topsim} and generalisation, this is surprising since the higher degree of \textsc{topsim} should imply that the language is more structured. Moreover, both, $\textsc{rsa}_{si}$ and $\textsc{rsa}_{li}$ have not drifted away from the image features. This combination, \textit{in theory}, should be ideal for discriminating image pairs from the Winoground dataset since it was designed to be discriminative with compositional visio-linguistic reasoning. However, in \textit{practice} this is not the case.

\section{Discussion}
In this work, we revisited the representational alignment problem in a common setup used in emergent communication and proposed a solution to this underrepresented problem. We corroborated earlier findings by showing that agents align their image representations and rely on spurious image features instead of human-like concept-level information \citep{bouchacourt-baroni-2018-agents}. We then showed that inter-agent alignment strongly correlates with the commonly used \textsc{topsim} metric. Our solution to the alignment problem involves an alignment penalty that forces the agents to remain aligned with the input features and mitigates the alignment problem without decreasing communicative success. Finally, when agents are tested on more challenging Winoground pairs we observed reasonable but lower performance regardless of whether image representations were similar to the image embeddings or not. With this work, we hope that the alignment problem will receive more attention in the field of emergent communication, as is already the case in adjacent fields \citep{sucholutsky2023getting}. 

\subsection{Importance of representational alignment}
It is common practice in simulations of emergent communication to process (visual) inputs into an agent-specific hidden representation and update their weights simultaneously \cite[e.g.][]{Lazaridou2017Multi-agentLanguage, bouchacourt-baroni-2018-agents, Chaabouni2019anti, chaabouni-etal-2020-compositionality, rita2022emergent}. As such, inter-agent alignment, \textit{irrespective of the input form}, likely happens in other simulations too. This phenomenon is therefore potentially widespread and perhaps the cause for findings that are at odds with experimental findings. While it is not always the case that the representation structure we \textit{expect} to help solve a task will do so \cite[e.g.][]{montero2021the, Xu2022Compositional}, such discrepancies may hinder the use of emergent communication models in developing a more natural understanding of human languages and leave them less suitable for directly simulating language evolution phenomena. Especially if we want machine representations of natural language to align with human representations \citep{sucholutsky2023getting}. \textsc{rsa} should therefore be used to rule out, or at the bare minimum report about, representational alignment in the future. 

\subsection{\textsc{topsim} and representational alignment}
Measuring representational alignment using \textsc{rsa} is similar to how \textsc{topsim} measures the structure in messages. They differ in their inputs but both calculate the Spearman-ranked correlation between metric-agnostic pairwise distances. Crucially, the input makes all the difference, the inputs for \textsc{rsa} are from both agents and are trained independently, whilst \textsc{topsim} only assesses the relation between the fixed inputs and learned output. Despite the similarities, the metrics thus describe different phenomena and are rarely reported simultaneously.

We hypothesise that the relationship between \textsc{topsim} and inter-agent representational alignment is a by-product of the setup, which in essence implies that the listener has to align its representation $r_{l}$ to the speaker representation $r_{s}$ \citep{rita2022emergent}. It has to do so using only the speakers' messages, being an abstraction of $r_{s}$. A solution to this problem is to align representations, which eases the listeners' training objective. If the speaker consistently produces structured messages during training, aligning $r_{l}$ to $r_{s}$ is easier, thereby causing higher inter-agent alignment. 
Essentially, this renders \textsc{topsim} to be an \textit{indirect} metric for the rate of alignment, for which $\textsc{rsa}_{sl}$ is a \textit{direct} metric. In the context of learnability, the relationship between \textsc{topsim} and inter-agent alignment and the fact that alignment always occurs can be seen as reasons for why languages with higher \textsc{topsim} are easier to learn \citep{LiBowling2019Ease-of-Teaching, cheng-etal-2023-correspondence}. This underscores the need to report inter-agent representational alignment to avoid conclusions drawn about the effect of specific interventions on \textsc{topsim} which may be attributable to inter-agent alignment. 

\subsection{Targeted o.o.d. evaluations}
An important implication of our findings concerns the standard practice of reporting o.o.d. accuracy where the agents are tested on unseen input after training \cite[e.g.][]{auersperger-pecina-2022-defending, conklin2023compositionality}. This should inform about the agents' ability to generalise from one dataset (e.g., MS COCO) to another dataset (e.g., the Winoground pairs) much like human language allows us to talk about an infinite number of situations. Crucially, this overlooks the representational alignment problem in that we do not know \textit{what} the agents are precisely generalising about. This problem can be mitigated with the alignment penalty to assess generalisation more directly or at least should be taken into consideration. 

We assess o.o.d. performance on the more challenging Winoground pairs as a proxy for the agents' ability to endow in compositional reasoning for image-based settings. Good performance on the Winoground dataset requires a grounded language that can be used to create compositional messages since the objects and their underlying relations need to be described. In general, we suggest to start evaluating simulations of referential games on targeted strict tasks, like probing state-of-the-art vision language models on e.g., visio-compositional \citep{Thrush2022Winoground, diwan-etal-2022-winoground, Hsieh2023SugarCrepe, Arijit2023Cola} or spatial \citep{kamath-etal-2023-whats} reasoning. Re-purposing such datasets can reveal more directly whether agents develop the attested communicative abilities that are trivial to humans without having to rely on metrics. Our results illustrate this through a shortcoming of the \textsc{topsim} metric. We observed that agents still struggle with distinguishing pairs of \textit{conceptually similar} Winoground images even though \textsc{topsim} is higher with the alignment penalty. If the language protocol were to communicate concept-level information \textit{and} compositional messages were created, we should not observe this struggle, meaning that the emerged protocols do not enable human-like communicative success. 

Interestingly, the o.o.d. performance remains substantially above chance in the $ce+\textsc{rsa}$ setting. Given that MS COCO is not a dataset for learning to model compositionality, this delineates the limits of what can be achieved qua performance based on MS COCO image features in the Winoground context. Nevertheless, this leaves open the question of above-chance performance on Gaussian noise with the $ce+\textsc{rsa}$ loss. A tentative explanation is that the higher inter-agent alignment on noise input (${M_{ce}=.428}$, ${M_{ce+\textsc{rsa}_{sl}}=.543}$, ${t=-8.71}$, ${p<.001}$) alleviates part of the problem (Figure \ref{fig:rsa_sl}). To validate this, future experiments should involve controlling the prior distributions of the agents' image encoders by training their vision modules on different data. Doing so ensures that they have to communicate about novel objects and cannot rely on similar representations.

\section{Conclusion}
This paper revisits the underrepresented alignment problem present in the referential game often used in simulations of emergent communication. Specifically, we focused on the problem of increasing alignment between agent-image representations in combination with a decreasing alignment between the input and agent representations. We first confirmed that agents align their image representations while losing connection to their input, meaning that the emergent languages do not appear to encode human-like visual features. We then showed that, in the common setup, inter-agent alignment is related to topographic similarity, and argued that this renders \textsc{topsim} an \textit{indirect} metric of the rate of inter-agent alignment. To further investigate the effects of alignment, we introduced an alignment penalty to mitigate the alignment problem and showed that the communicative ability on a strict compositionality benchmark did not improve, leaving the question of inducing compositional generalisation in emergent communication for images unsolved. Our findings underscore the need to better understand the divergence between human and artificial language emergence within the prevalent referential setup and highlight the importance and potential impact of representational alignment. We hope that future work rules out or at least reports about representational alignment.  

\section{Limitations}
Our work has a few notable limitations. First, it only involves the referential game. Another popular variant, the reconstruction game \cite[e.g.][]{Chaabouni2019anti, chaabouni-etal-2020-compositionality, lian-etal-2021-effect, conklin2023compositionality}, requires the listener to reconstruct the input object based on the speakers' message. Since this setup has a different objective and presents different learning biases, it may have different results. We still expect the results to be similar as there is no pressure to retain alignment between the image input and agent representation. It would, however, be interesting to investigate whether the language protocol in this scenario is more structured than in the referential game. 

Another limitation in our setup is that we only consider the scenario with two agents, which may be a requirement for alignment to be possible. Since experiments with human participants show that larger communities create more systematic languages \citep{raviv2019larger}, simulations on emergent multi-agent communication with populations of agents are also conducted, but these yield mixed results. The emergent communication protocols oftentimes do not evolve to be more structured unless explicit pressures such as population diversity or emulation mechanisms are introduced \cite{rita2022emergent, chaabouni2022emergent}. \citet{michel2023revisiting} however, showed that population setups can result in more compositional languages if agent pairs are trained in a partitioned manner to prevent co-adaptation. Despite the mixed results, we believe that emergent communication with populations of agents is ecologically more valid and could result in different alignment effects. Much like how \citet{tieleman2019shaping} showed that autoencoders encode better concept category representations when they learn representations in a community-based setting with multiple encoders and decoders collectively.

The final limitation of our study regards its scale. While simulations of emergent communication are typically conducted on relatively small-scale datasets, human language emergence is accompanied by rich and diverse multi-modal experiences. Recent results in the field of computer vision suggest that dataset diversity and scale are the primary drivers of alignment to human representations \citep{Conwell2023whatcan, muttenthaler2023human}. As such, this key difference between the setting of artificial emergent communication and human language emergence can drive the observed differences in representations. Due to the difficulty of interpreting these representations, we see this as another reason to evaluate emergent protocols on more strict datasets with clear pragmatic value for humans.

\bibliography{anthology, custom, EmeCom}
\bibstyle{acl_natbib}

\appendix

\section{Channel capacity}
\label{sec:Appendix-capacity}
To test to what degree communicative success, \textsc{topsim}, and representational alignment are confounded with the communication channel capacity, we ran simulations altering the vocabulary size ($V =\{3,5,10,20,40,50,100\}$) and message length ($L=\{2,3,5,10,50,100\}$) resulting in 42 parameter settings per loss type.

\begin{figure}[ht]
    \centering
    \includegraphics[width=1\linewidth]{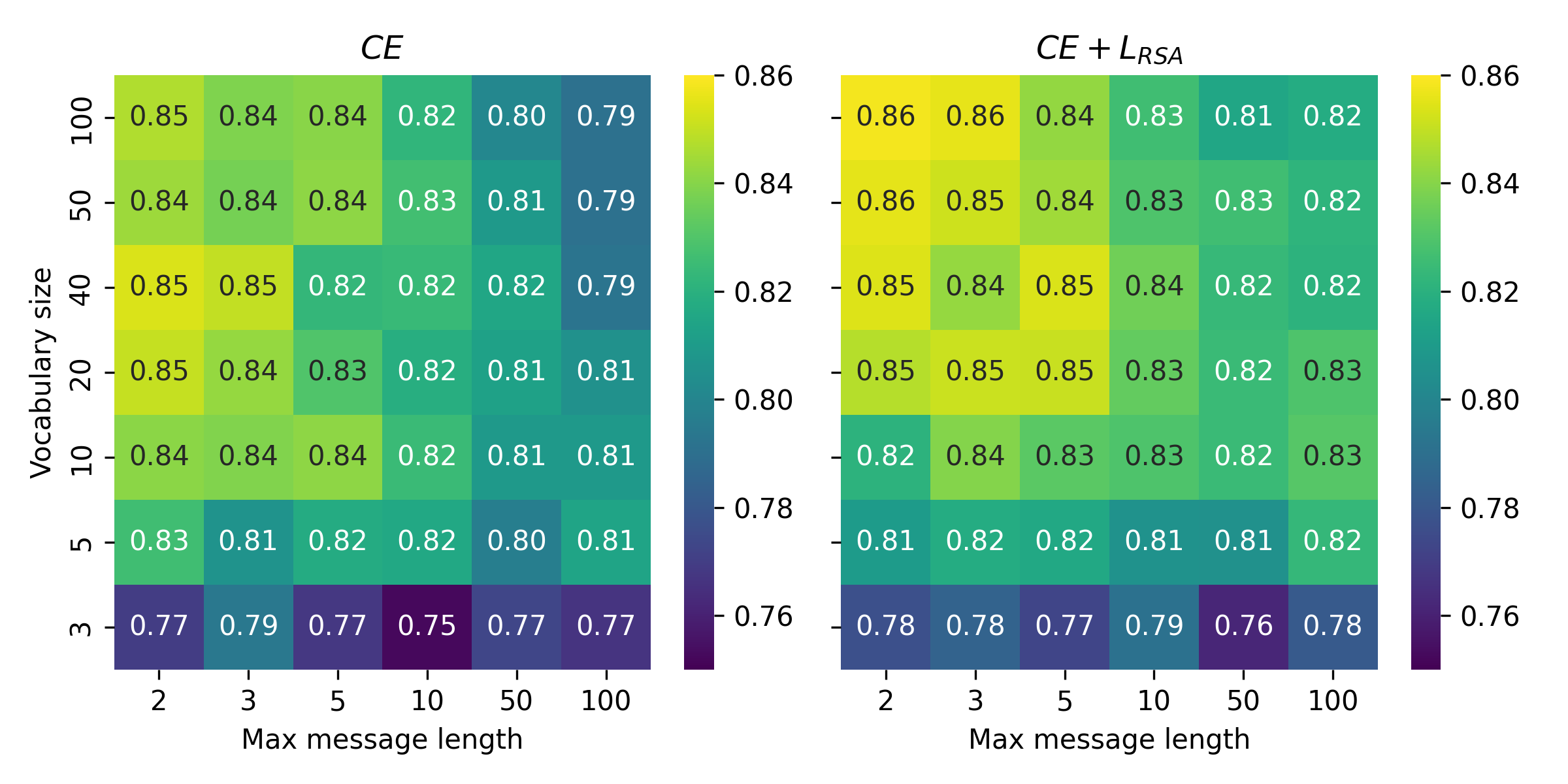}
    \caption{The validation accuracy as a dependent factor of the vocabulary size and maximum message length. Values are averages across 15 seeds.}
    \label{fig:heatmap_acc}
\end{figure}

\begin{figure}[ht]
    \centering
    \includegraphics[width=1\linewidth]{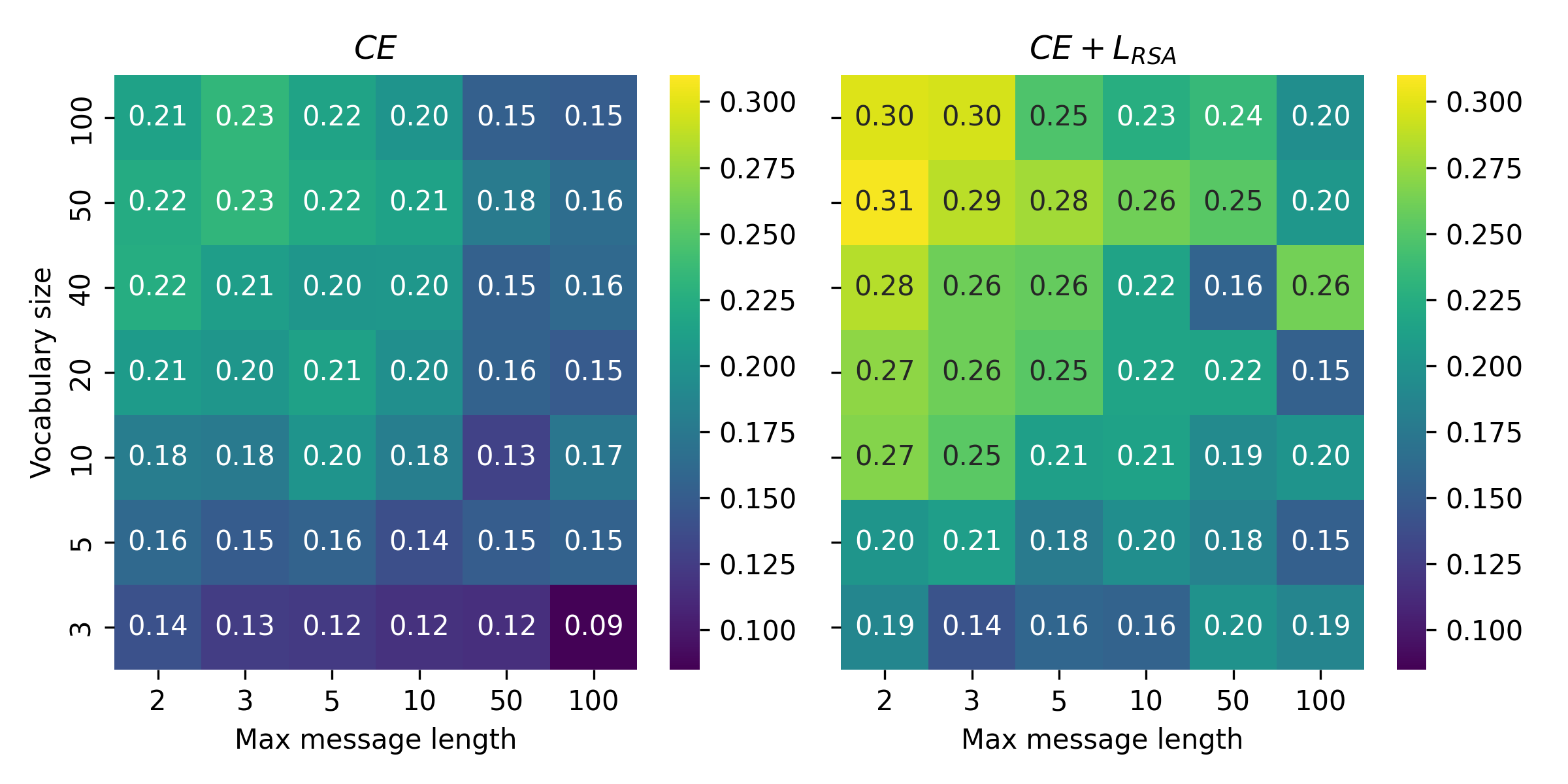}
    \caption{\textsc{topsim} as a dependent factor of the vocabulary size and maximum message length. Values are averages across 15 seeds.}
    \label{fig:heatmap_topsim}
\end{figure}

Overall, performance is relatively independent of the chosen configuration, but vocabulary size influences success more than message length (Figure \ref{fig:heatmap_acc}). The hyperparameters that resulted in the best validation accuracy \cite[i.e., generalisation;][]{chaabouni2022emergent} for the standard $ce$ setup were $V = 40$ and $L = 2$. These parameters are used to produce the results in the main paper. Contra expectations, the vocabulary size also influenced \textsc{topsim} more than message length. It, especially in the case of $ce+L_{\textsc{rsa}}$, is higher when messages are shorter but have access to a larger vocabulary (Figure \ref{fig:heatmap_topsim}).

\begin{figure*}[ht]
    \centering
    \includegraphics[width=1\linewidth]{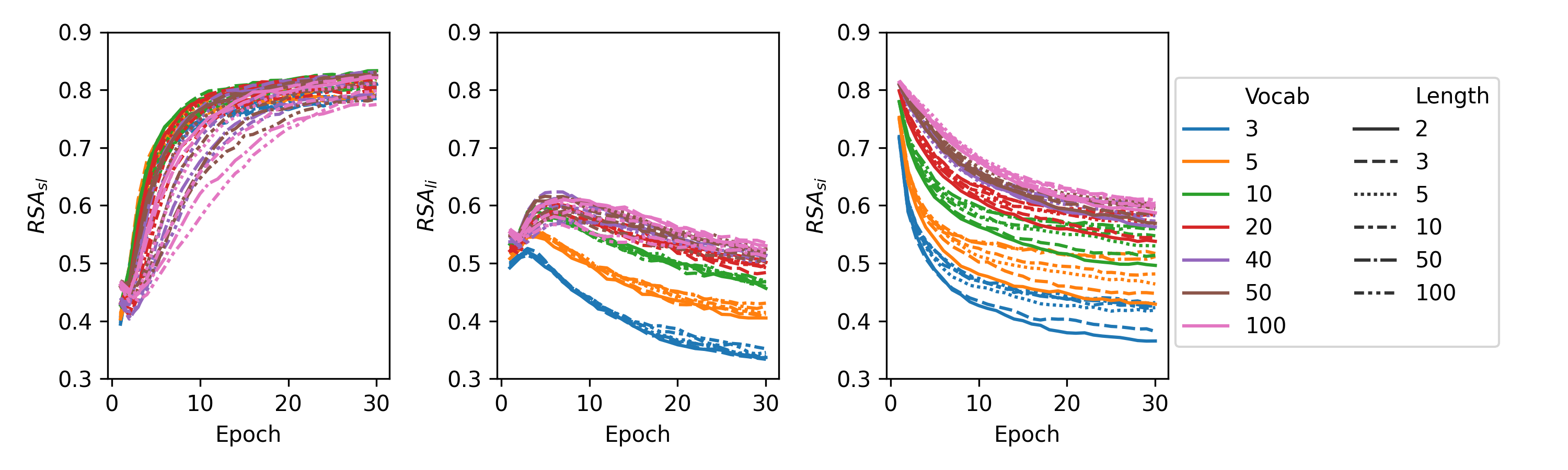}
    \caption{Representational alignment metrics averaged over 15 simulations with the standard $ce$ loss. Regardless of channel capacity, representational alignment always occurs while losing relation to the input.}
    \label{fig:RSA_all}
\end{figure*}

Figure \ref{fig:RSA_all} shows that, regardless of capacity, inter-agent alignment ($\textsc{rsa}_{sl}$) increases while image-agent alignment ($\textsc{rsa}_{si}$ and $\textsc{rsa}_{li}$) decreases with the $ce$ loss. Interestingly, $\textsc{rsa}_{sl}$ is agnostic to capacity but a larger vocabulary size, not message length, reduces the degree of drifting away from the input. We hypothesise this to result from lower pressure to compress rich continuous embeddings into smaller discrete vocabulary embeddings.

\section{Best hyperparameters}
\label{sec:hyperparams}
The parameters used to run the experiments in the main paper were the following:

\begin{table}[ht]
\centering
\begin{tabular}{l|l}
\textbf{Parameter}          & \textbf{Value} \\\hline
Batch size                  & 32 \\ 
Optimiser                   & Adam \\
Learning Rate (S \& L)      & 0.01 \& 0.001 \\
Vocabulary size ($V$)       & 40 \\
Message length ($L$)        & 2 \\
Hidden size (S \& L)        & 768 \& 768 \\
Embedding size              & 50 \\
Listener cosine temperature & 0.1 \\
Seeds                       & \begin{tabular}[c]{@{}c@{}}16,22,41,56,67,\\77,14,78,99,23,\\82,40,51,37,62\end{tabular} \\
\end{tabular}
\caption{Best-performing parameters resulting from the parameter sweep.}
\label{tab:best_parameters}
\end{table}

\section{Interaction of the alignment term on the cross-entropy loss}
\label{sec:interaction_ce_ce_aux}
\begin{figure}[ht!]
    \centering
    \includegraphics[width=1\linewidth]{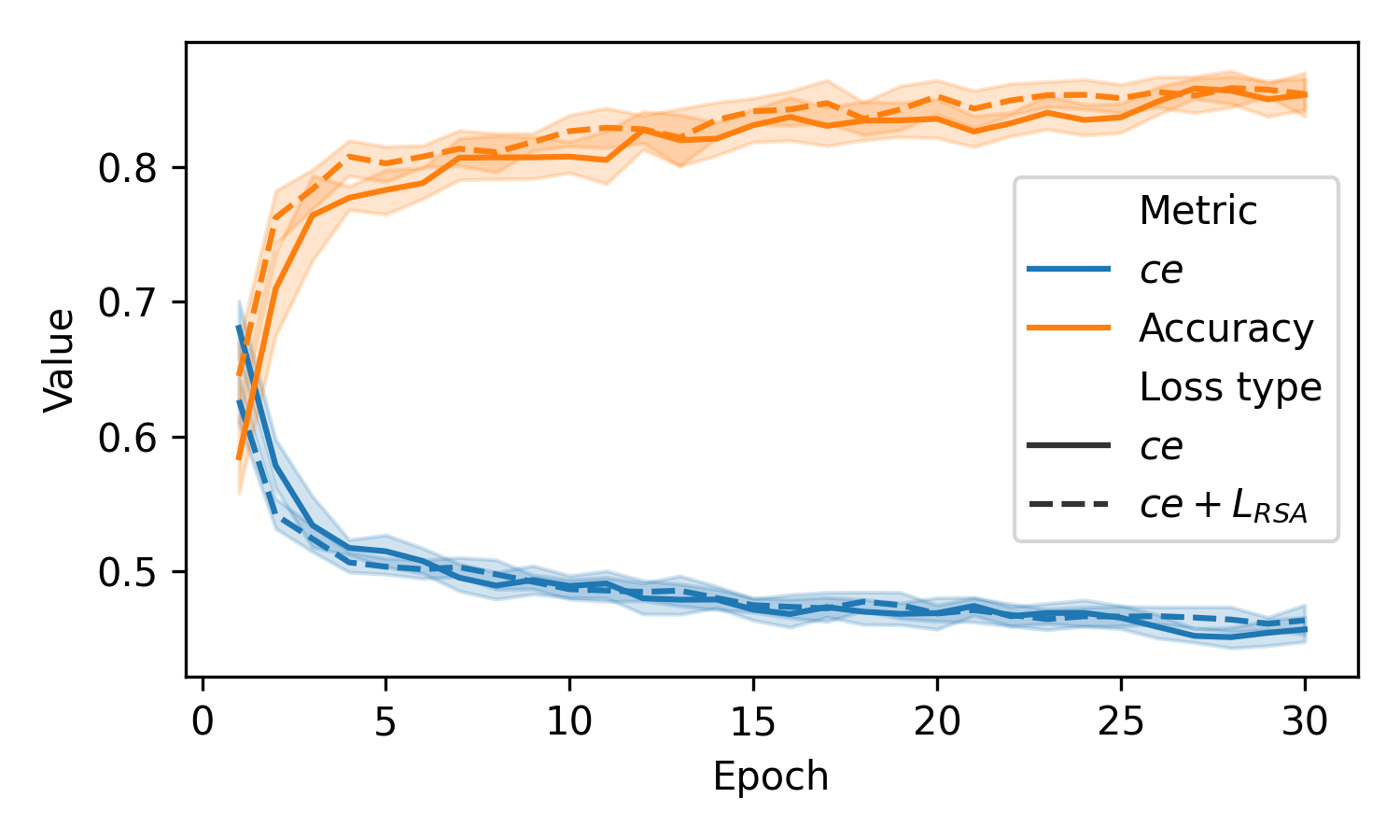}
    \caption{Learning curves (accuracy) and cross-entropy loss ($ce$) for both loss settings. There is virtually no effect of the auxiliary term $L_\textsc{rsa}$ on the cross entropy loss or communicative success.}
    \label{fig:ce_validation}
\end{figure}

To ensure that there is no impact of the alignment penalty on the pressure for communicative success, we ablated the $L_{\textsc{rsa}}$ term of our proposed loss function and found that both, communicative success and $ce$ are not affected by the alignment penalty (Figure \ref{fig:ce_validation}). Corroborating that only the $ce$ term provides pressure for successful communication (§\ref{sec:alignment_mitigated}).

\section{Pre-trained vision modules}
\label{sec:vision_modules}
Although it is in principle possible to train the vision module of the agents from scratch \citep{Roberto2021Interpretable}, in our work, agents' perception stems from a pre-trained vision-language model. Although there is reason to believe that DinoV2 embeddings capture high-level, conceptual image features useful for discriminating image pairs \cite{oquab2024dinov}, we assessed the degree to which the alignment problem occurs for different pre-trained models despite encoding the same objects. We ran additional simulations using image features obtained from ResNet \citep{he2016deep} and CLIP \citep{radford2021learning} for 6 different parameter settings with the $ce$ loss function. Here we used the parameters that resulted in the best, worst, mean, and quantile validation performance from the parameter sweep in appendix \ref{sec:Appendix-capacity} (see Table \ref{tab:parameters}), and a sensible setup with $V = 10$ and $L = 5$.

\begin{table}[ht]
\centering
\begin{tabular}{cc||c}
\textbf{Msg. Length ($L$)} & \textbf{Vocab. Size ($V$)} & \multicolumn{1}{l}{\textbf{Vision}}                                             \\ \hline
2                       & 40                      & \multirow{6}{*}{\begin{tabular}[c]{@{}c@{}}DinoV2\\ CLIP\\ ResNet\end{tabular}} \\
3                       & 10                      &                                                                                 \\
5                       & 5                       &                                                                                 \\
5                       & 10                      &                                                                                 \\
10                      & 3                       &                                                                                 \\
50                      & 100                     &                                                                                
\end{tabular}
\caption{The parameters for running additional simulations with CLIP and ResNet to assess the robustness of our results. Each combination was run for 15 different seeds. Note: results for the DinoV2 simulations are from the sweep.}
\label{tab:parameters}
\end{table}

\begin{figure}[!ht]
    \centering
    \includegraphics[width=1\linewidth]{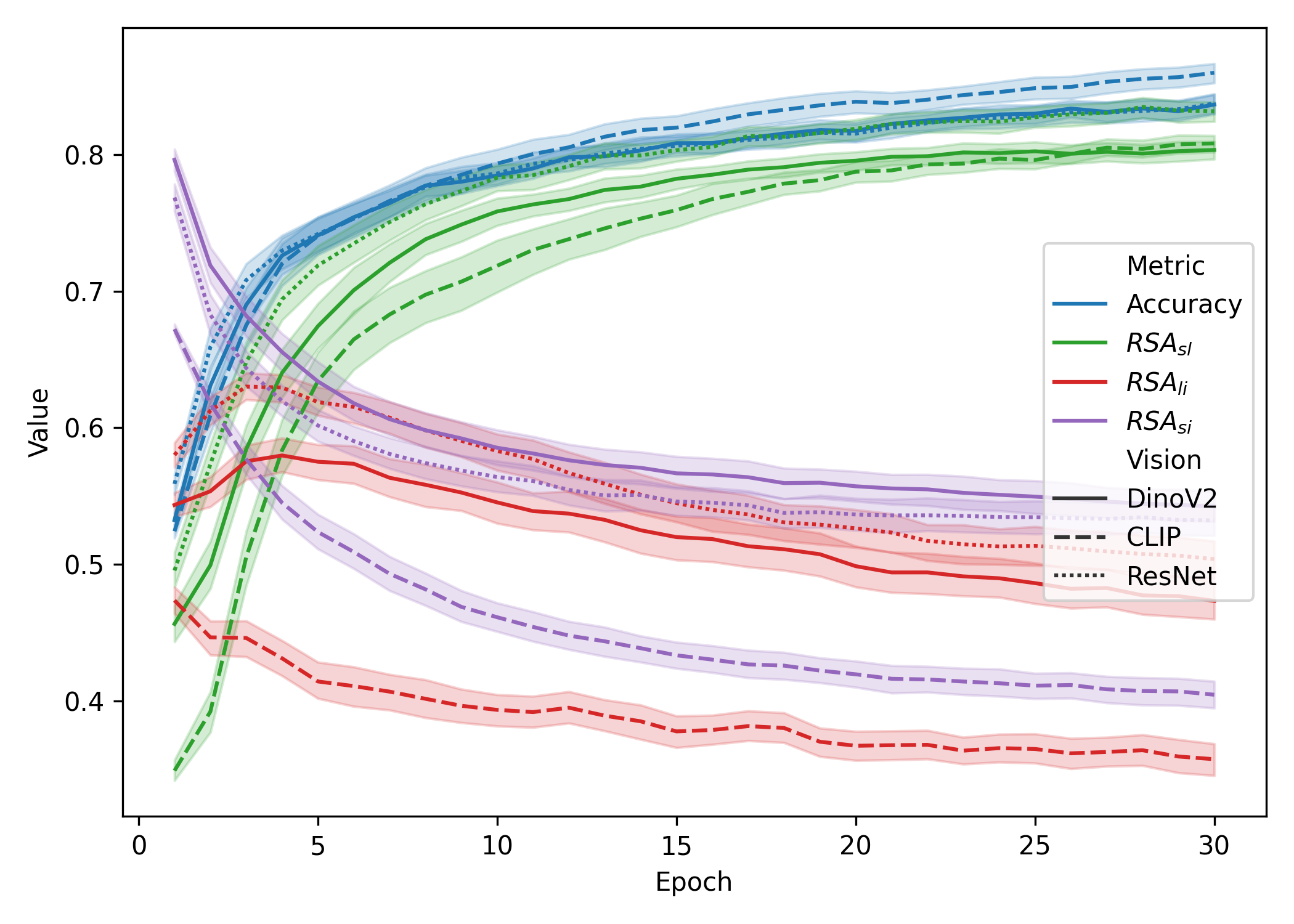}
    \caption{Learning curves (accuracy) and \textsc{rsa} metrics for different vision models averaged over 6 parameter settings with 15 seeds each. The representational alignment problem always occurs. Line style corresponds to the vision module used to obtain image embeddings and colour indicates the metric. Areas indicate the 95\% confidence intervals.}
    \label{fig:vision}
\end{figure}

Figure \ref{fig:vision} shows clearly that inter-agent alignment \textit{increases} while agent-image alignment \textit{decreases} for all models. In addition to the similar results reported by \citet{bouchacourt-baroni-2018-agents} for VGG ConvNet embeddings, both 4096 and 1000 layers, we can confirm that the problem is agnostic to the input embeddings. Interestingly, agent representations drift most for CLIP embeddings. Nevertheless, the agents still develop a successful communication strategy, indicating that out-of-the-box CLIP embeddings are the least useful for agents in finding a (non-grounded) solution. No such differences are seen when the agents are trained with the additional alignment penalty term, inter-agent and image-agent alignment remain high for all models.

\end{document}